# Informative Path Planning to Explore and Map Unknown Planetary Surfaces with Gaussian Processes

Ashten Akemoto
University of Hawaii
1680 East-West Road.
Honolulu, HI 96822
aakemoto@hawaii.edu

Frances Zhu
University of Hawaii
1680 East-West Road.
Honolulu, HI 96822
zhuf@hawaii.edu

*Abstract*— Some environments remain largely unexplored, like unvisited planetary surfaces or regions of the ocean. Many of the destinations we are interested in have never been explored and thus, there is no a priori knowledge to leverage. Instead, a vehicle must sample upon arrival, process this data, and either send this information back to a teleoperator or autonomously decide where to go next. Teleoperation is suboptimal in that human intuition can be imprecise and cannot be mathematically guaranteed to yield an optimal result. Given a surface environment, a mobile agent will map the distribution of a scalar variable without any prior information, with a degree of confidence of model convergence, and while minimizing distance traveled. Science-blind approaches to covering a surface area include a predefined path with waypoints for a vehicle to locomote in an "open-loop" policy, like the Boustrephedon or Spiral patterns. Information theoretic approaches iteratively gather information to feed back into a model or policy, which then updates with a waypoint to visit next. Gaussian processes have been popularly incorporated into active learning exploration strategies due to their ability to incorporate information into a parameter-free model and quantify a confidence metric with every model prediction. We evaluate the performance of this informative path planning algorithm in mapping an environment on three different surfaces: parabola, Townsend, and hydration value across a lunar crater from LAMP data. The Gaussian process model was tasked to learn these relationships across these specific surfaces to investigate the efficacy of the learner and policy on a noiseless vs noisy surface, a convex vs a nonconvex surface, and a smooth well-behaved function vs an unknown real function. We quantify model variance, model root-mean-squared error, distance, and surface global minimum identification of the various methods in exploring a range of surfaces with no a priori knowledge. The results show that the information-driven methods significantly outperform naive exploration methods in minimizing model error and distance with potential of convergence.

## TABLE OF CONTENTS

## 1. INTRODUCTION

Some environments remain largely unexplored, like unvisited planetary surfaces or regions of the ocean. Aspects of these environments are very difficult to sense without sampling in situ, like the amount of water-ice underneath the regolith surface layer or thermophysical attributes in deep water. Many of the destinations we are interested in have never been explored and thus, there is no a priori knowledge to leverage. Instead, a vehicle must sample upon arrival, process this data, and either send this information back to a teleoperator or autonomously decide where to go next. Teleoperation is suboptimal in that human intuition can be imprecise and cannot be mathematically guaranteed to yield an optimal result.

As these destinations of interest also limit communication with a human operator, autonomy is imposed onto the intelligence framework. Missions may incorporate autonomy at different levels, ranging from incorporating the machine learning predictions from the model as a suggestion at every decision point or necessitating low-level autonomy for short bursts during dives underwater but still relying on an operator upon resurfacing. Teleoperators can spend quite a bit of time making a decision. Coordinating a decision amongst other operators compounds in complexity and time. This paper will investigate full autonomy with absolutely no human intervention in the exploration process to offer science teleoperators an intelligent suggestion system or to entirely replace the scientist. Surface operations can be formalized into a cost-function based exploration algorithm that is precise, provides mathematical guarantees, and implemented in real or near-real time. Reducing teleoperations and promoting autonomy reduces personnel effort and cost, reduces mission length and risk, and offers verifiable results.

We assume that a vehicle arrives at a destination of interest, like the Perseverance rover at the Martian Jezero Crater, with the intention to comprehensively map the environment with little to no a priori knowledge. The mapping activity is a regression problem, relating a continuous scalar to a location on the environment surface or volume. We want to train a model to accurately predict the scalar value of the variable of interest distributed across the surface volume based only on the iteratively gathered samples, while minimizing distance traveled to gather such samples and quantifying model convergence. This motivating problem formulation draws from space and underwater exploration applications, where little is known about the environment, total mission duration is limited, and sampling is expensive.

Exploration strategies range from human teleoperation to autonomous robotic operations, which range from science-blind to information theoretic based methods. Science-blind approaches to covering a surface area include a predefined path with waypoints for a vehicle to locomote in an "open-loop" policy, like the Boustrephedon or Spiral patterns. Information theoretic approaches iteratively gather information to feed back into a model or policy, which then updates with a waypoint to visit next. Gaussian processes have been popularly incorporated into active learning exploration strategies due to their ability to incorporate information into a parameter-free model and quantify a confidence metric with every model prediction. Current information-driven strategies in autonomous robot exploration provide myopic (or short-sighted) methods [1], map only scalar variables for underwater environments [2], or require human scientist prior belief or bias [3]. The most relevant information-driven planetary exploration research compositionally maps rocky environments for rover vehicles with the combination of remote and in-situ spectroscopy [4].

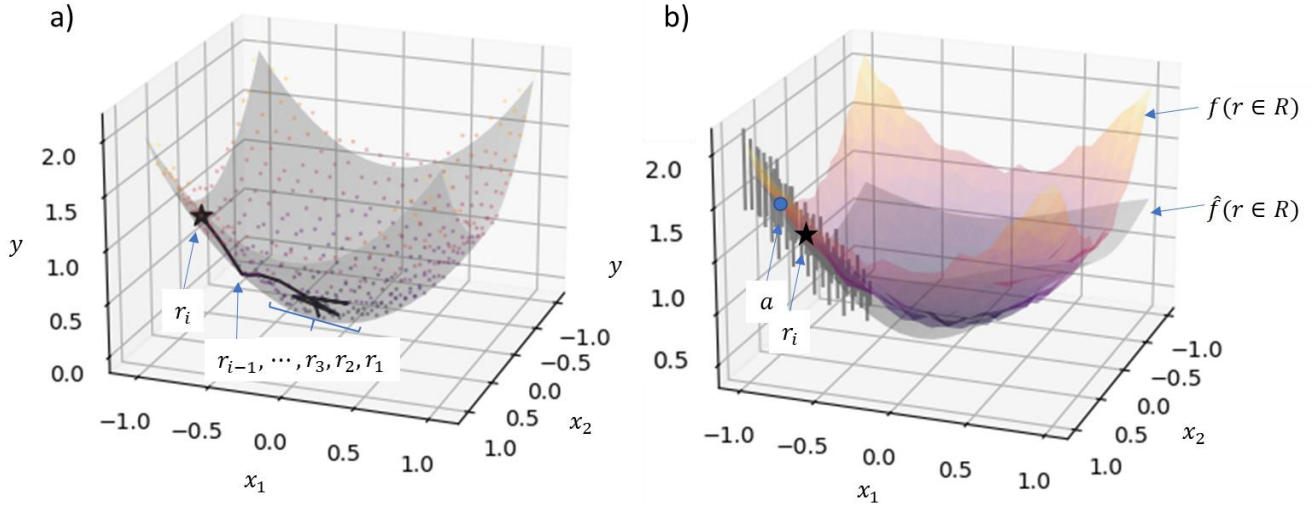

Figure 1: a) Traverse path of the rover along an environment surface, b) True environment surface and model predicted surface

## 2. PROBLEM FORMULATION

Given a surface environment ($\mathbb{R}^2$ space), a mobile agent will map the distribution of a scalar variable without any prior information, with a degree of confidence of model convergence, and while minimizing distance traveled. The objective function, $J$, aims to minimize a learned model's prediction $\hat{f}(r, a)$ with respect to ground truth $f(r, a)$ at a location on the surface $r$ across a set of locations $R$ and proposed action $a$, graphically depicted in Figure 1.

$$J^* = \min_a \left| f(r,a) - \hat{f}(r,a) \right|_2, r \in R$$

This model error takes the form of the $L_2$ norm or root mean-squared (RMS) error. The product of optimizing this objective function is the action, a proposed waypoint to sample at next. The mobile agent will follow the iterative process of 1) sample the surface environment at location $r_i$, 2) form a representative model of the surface given historical measurements $\hat{f}(r_1, \dots, r_i)$, 3) decide upon the next waypoint $a$ to visit, and repeat this iteration until the model error is sufficiently small and/or the model has indications of convergence.

## 3. METHODOLOGY

This paper studies two different classes of methods: science blind and information driven. Science-blind methods do not incorporate *a priori* knowledge or iteratively collected measurements of the subject whereas information-driven methods do incorporate measurements as updates to some policy to update the next action. We baseline the Boustrephedon and Spiral coverage path planning algorithms, seen in Figure 2 and replicated from Laine and Kazuya [1].

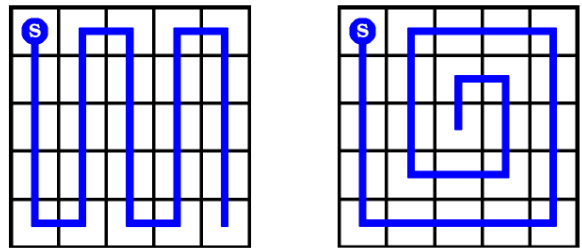

(a) Boustrophedon motion.   (b) Spiral motion.

Figure 2: Representation of the two basic motion patterns for minimally complex coverage path planning algorithms [1]

The hypothesis is that an information-driven method, specifically a Gaussian process model, efficiently explores a space in quickly reducing model error in a minimal number of samples and distance. Intuitively, the Gaussian process exploration strategy seeks out locations that the underlying Gaussian process model knows little about (low confidence or high variance) and prioritizes those surprising locations to explore. Step by step, the information-driven exploration algorithm procedure is:

1) Set hyperparameters and constraints
   a. total environment surface area $R$
   b. distance horizon constraint in selecting the next waypoint $r_{way}$
   c. prediction horizon $r_{pred}$
   d. covariance function $k(\cdot)$, a radial basis function
   e. training iterations
   f. size of initial training set $n_0$
2) Collect initial random data with a random walk
   a. Initialize the rover's location to a single location on the surface. This location constitutes the first input $\vec{x}_1$ of the training set.
   b. Sample at this location. This scalar constitutes the first output or target $y_1$ of the training set
   c. Traverse to a random nearest neighbor location (8 potential neighbors) while avoiding revisits
   d. Return to step 2.b to append more data to the training set for a total of $n_0$ input-output pairs
3) Explore the surface $R$ until the Gaussian process model converges
   a. Check if the Gaussian process model reaches a threshold prediction variance across the surface $\hat{V}_R$ or has sampled a threshold number of unique data points $n_{max}$
   b. Train the Gaussian process model on the $n$ input-output pairs in the training set thus far $(X, Y) \to \hat{f}$ where $X = \begin{bmatrix} \vec{x}_1 \\ \vdots \\ \vec{x}_n \end{bmatrix}$ and $Y = \begin{bmatrix} y_1 \\ \vdots \\ y_n \end{bmatrix}$
   c. Predict scalar expected values $\hat{Y}_{pred}$ and variance $\hat{V}_{pred}$ in the prediction horizon $r_{pred}$
   d. Generate a control policy $g^*$ that identifies the location in the prediction horizon with the highest variance $r_{Vmax}$
   $$r_{Vmax} = g^* = \underset{r \in r_{pred}}{\mathrm{argmax}}\; \hat{V}_{pred}$$
   e. Traverse to the nearest neighbor location in the direction of this high-variance location $r_{Vmax}$. The action $a$ is the next location
   $$a = |r_{way} - r_{Vmax}|_2$$
   f. Sample the value $y_{n+1}$ at this location $x_{n+1}$ and append to the training set

We evaluate the performance of this informative path planning algorithm in mapping an environment on three different surfaces: parabola, Townsend, and hydration value across a lunar crater from LAMP data. These surfaces have two independent dimensions (planar position $r = (x_1, x_2)$) and a third dependent dimension $y$; the dependent variable's algebraic relationship to position is known for the parabola and Townsend benchmark surfaces but unknown for the lunar ice data. The parabola surface is defined by:

$$y = x_1^2 + x_2^2 + \sigma_{noise}^2$$

where $\sigma_{noise}^2 = 0.02$ or $0$, $x_1 \in [-1: 0.1: 1]$, and $x_2 \in [-1: 0.1: 1]$. The Townsend surface is defined by:

$$y = -\left(\cos\big((x_1 - 0.1)x_2\big)\right)^2 - x_1 \sin(3x_1 + x_2) + \sigma_{noise}^2$$

where $\sigma_{noise}^2 = 0.02$ or $0$, $x_1 \in [-2.5: 0.1: 2.5]$, and $x_2 \in [-2.5: 0.1: 2.5]$.

The LAMP data contained a digital elevation map (DEM) $(r = (x_1, x_2, x_3))$ of the lunar South pole in 5 m spatial resolution and hydroxol data $y$ in 250 m spatial resolution, which was mostly present but significant swaths of data are missing near the crater rim, as can be seen in Figure 3.

The Gaussian process model was tasked to learn these relationships across all exploration strategies on these specific surfaces to investigate the efficacy of exploration policy on a noiseless vs noisy surface, a convex vs a nonconvex surface, and a smooth well-behaved function vs an unknown real function that was potentially discontinuous.

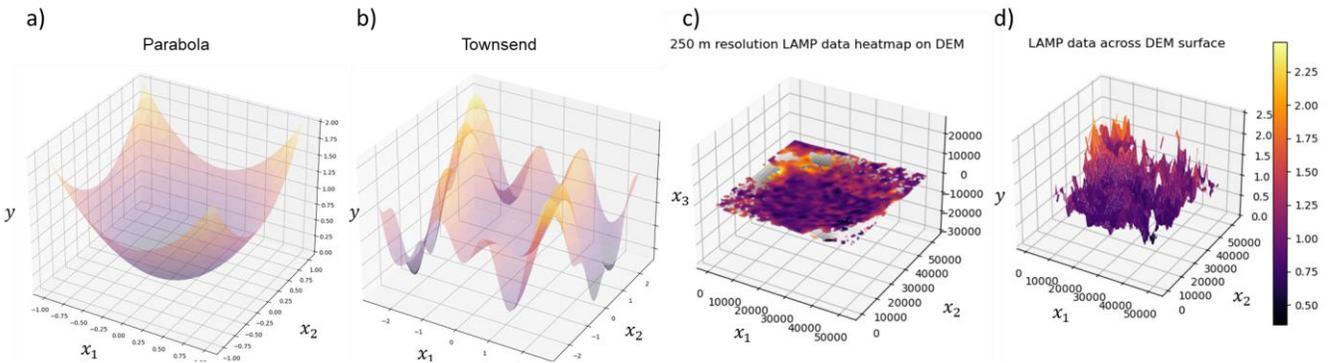

Figure 3: Surface environments to quantify exploration and learning strategies. a) parabola surface, b) Townsend surface, c) lunar South pole digital elevation map (DEM), d) LAMP data across DEM

The simulation and algorithm are encoded in Python with the source code, LAMP data, and simulation videos published on github (https://github.com/frankiezoo/GaussianProcessLunarExploration). We heavily utilize the GPyTorch code package for the exploration strategy [5].

## 4. EVALUATION METRICS

Metrics of performance in evaluating the science-blind and information-driven exploration methods include:

- RMS error upon Convergence
- Samples until Convergence
- Distance until Convergence
- Initial Global RMS / Final Global RMS error
- Error in Identifying Global Minimum

RMS error upon convergence and samples / distance until convergence are only relevant for trials that seem to asymptotically approach a constant value, which can only be speculated is this terminal value is unknown. To determine the concept of convergence loosely, the 2% settling time from control theory was adopted. The global RMS error between the model prediction and true values is inspected to verify that 1) there are enough data points to confirm convergence and 2) that the final values of RMS error stay within a 2% band of the final value $e_f$. The 2% error band $\Delta e_{2\%}$ is found by differencing the initial RMS error $e_0$ and final RMS error $e_f$. The index of convergence or samples until convergence $i_{2\%}$ is then found through minimizing an inequality comparing the error at an index $i$, $e_i$, and the upper limit of the 2% band.

$$\Delta e_{2\%} = 0.02(e_0 - e_f)$$

$$i_{2\%} = \underset{i}{\mathrm{argmin}}\, e_i \leq (e_f + \Delta e_{2\%})$$

A visual depiction of this calculation can be found in Figure 4. The distance $d_{i2\%}$ and RMS error upon convergence $e_{i2\%}$ can then be easily retrieved through a table lookup.

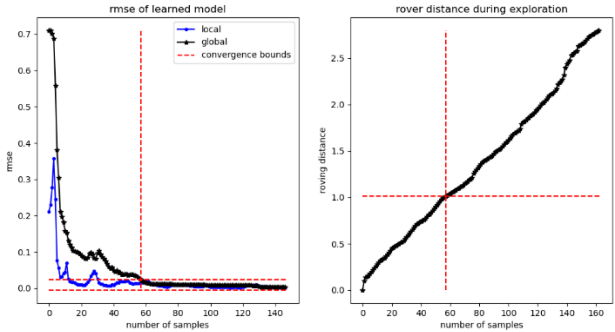

Figure 4: Finding the convergence RMS error value in an experiment trial and the subsequent roving distance

The initial global RMS / final global RMS error ratio compares the initial RMS error, which is globally evaluated with a blind rover, and the final RMS error, which is globally evaluated with a learned rover. This metric indicates the multitude of improvement to environment characterization and prediction offered by exploring the environment.

Finally, the error in identifying a global minimum is implicitly tied to the Gaussian process model's ability to accurately represent the true environment surface and the efficacy of the exploration strategy. Many applications not only want to characterize the environment, but are also motivated to seek extrema, like how a lunar South pole prospecting mission hopes to find a maximum water-ice reserve. The true minima for the benchmark and lunar surfaces are: parabola (0,0), Townsend (2.55, 0.026), lunar (1, 0.5, 0.034).

## 4. RESULTS

The spiral and Boustrephedon of varying waypoint horizons, and Gaussian process (GP) informed exploration strategies of various prediction horizons (Table 2) were evaluated on the three surfaces of varying surface size, toggling between a noiseless and noisy surface (Table 1).

Table 1: Range of simulated environments

| Environment Topography | Size of Surface | Noise Level |
|---|---|---|
| Parabola | $x_1 \in [-1: 0.1: 1]$ | Noiseless |
| | $x_2 \in [-1: 0.1: 1]$ | Noisy |
| Townsend | $x_1 \in [-2.5: 0.1: 2.5]$ | Noiseless |
| | $x_2 \in [-2.5: 0.1: 2.5]$ | Noisy |
| Lunar Crater LAMP data | $x_1 \in [-1.5: 0.25: 1.5]$ | Noisy |
| | $x_2 \in [-1.5: 0.25: 1.5]$ | |
| | $x_1 \in [-3: 0.25: 3]$ | Noisy |
| | $x_2 \in [-3: 0.25: 3]$ | |

Table 2: Range of exploration strategies and number of trials represented by results

| Exploration Strategy | Waypoint and Prediction Horizons | # of Trials Each |
|---|---|---|
| Spiral | $1\Delta x$ | 2 |
| | $2\Delta x$ | 2 |
| | $4\Delta x$ | 2 |
| Boustrephedon | $1\Delta x$ | 2 |
| | $2\Delta x$ | 2 |
| | $5\Delta x$ | 2 |
| | $10\Delta x$ | 2 |
| Gaussian Process Active Learner (GPAL) | $1\Delta x$ (nearest neighbor) | 6 |
| | $3\Delta x$ (local) | 6 |
| | $r \in R$ (global) | 6 |

The results comparing the three exploration strategies will be presented in order of ascending complexity:

- noiseless parabola,
- noisy parabola,
- noiseless Townsend,
- noisy Townsend,
- smaller 3km lunar crater swath, and
- large 6km lunar crater swath.

For all 2 x 2 subplot figures, the subplot components are a) rover traverse path, b) true environment surface vs predictions, c) variance and d) RMS error of local and global predictions vs. samples collected.

*Noiseless Parabola*

A noiseless, parabolic surface is a smooth, convex surface that should be straightforward to learn due to these properties. The total number of samples that could be sampled is 441.

All exploration strategies are able to converge. The spiral and Boustrephedon strategies result in identical models (no variance in box-whisker plots) as the model inputs are identical. The local and global GPAL strategies initialize randomly, which explains the variability. All exploration strategies dramatically improve the learned model's RMS error. All strategies can precisely locate the global minimum.

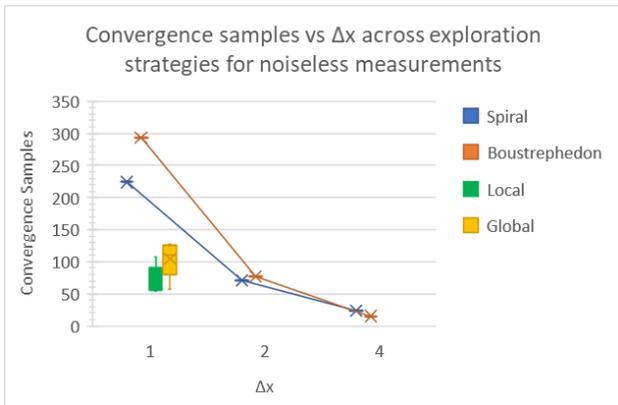

Figure 5: Comparison of samples until convergence across varying prediction and waypoint horizons on a noiseless parabola

The GPAL strategies converge at a similar number of samples as the spiral and Boustrephedon strategies sampling at $2\Delta x$, Figure 5. The local GPAL strategy needs between 54 – 107 samples to learn the surface well (12% - 24%).

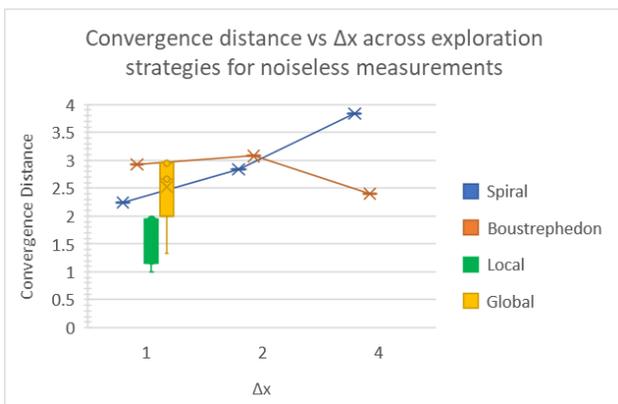

Figure 6: Comparison of distance until convergence across varying prediction and waypoint horizons on a noiseless parabola

The GPAL local strategy travels less distance (1 – 1.95 km) than any other strategy, with the global GPAL strategy in the range of the science-blind strategies, Figure 6.

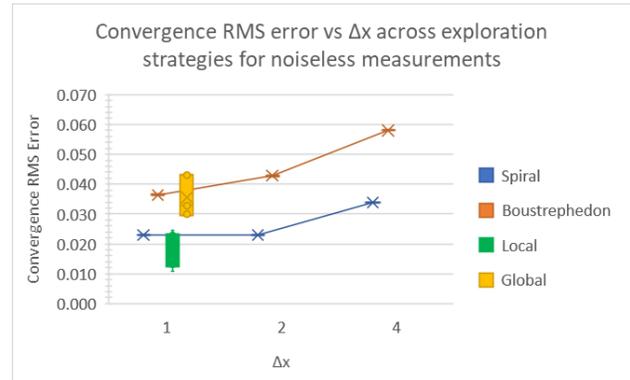

Figure 7: Comparison of RMS error until convergence across varying prediction and waypoint horizons on a noiseless parabola

The GPAL local strategy consistently converges to less RMS error (0.011 – 0.024) than any other strategy, Figure 7. The GPAL global strategy performed worse than the naïve spiral strategy but better than the Boustrephedon.

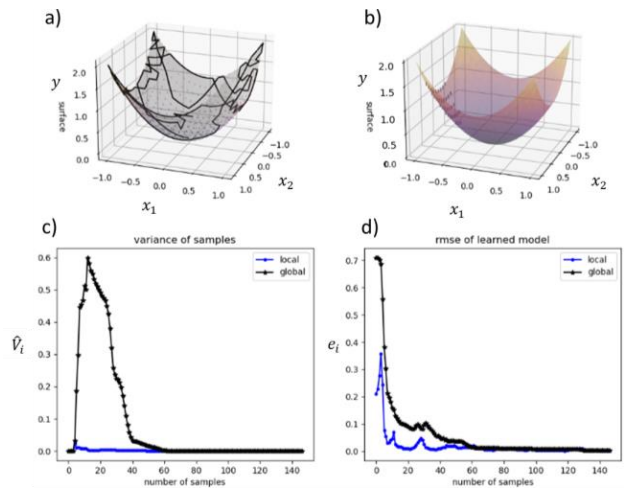

Figure 8: Example of GPAL local exploration strategy on noiseless parabolic surface.

Figure 8 shows the best strategy for a noiseless convex topography (GPAL local). The explorer correctly increases the variance of measurements, which intuitively decreases RMS error quickly. The variance then monotonically decreases. As for the path traversed, the explorer quickly travels to the edge of the area to constrain the boundaries. The global GPAL strategy could be underperforming because the global strategy generates waypoints that are too distracting from rigorously searching the local proximity.

*Noisy Parabola*

Unlike the noiseless case, the spiral and Boustrephedon strategies vary due to the noise in the training data. All exploration strategies moderately improve the learned model's RMS error. The addition of noise has increased the noise floor and achievable prediction error. All strategies can precisely locate the global minimum.

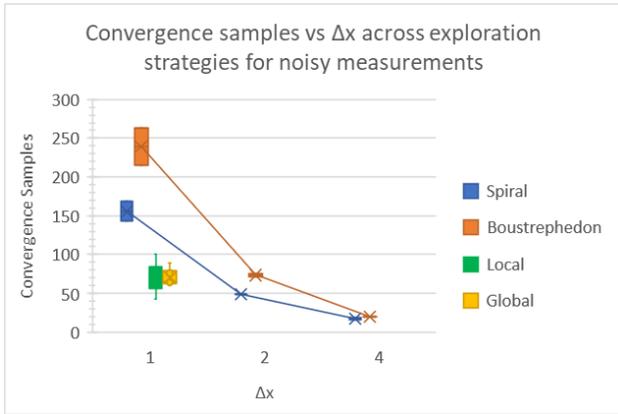

Figure 9: Comparison of samples until convergence across varying prediction and waypoint horizons on a noisy parabola

For a noisy convex surface, the GPAL strategies converge at significantly less samples than the naïve spiral and Boustrephedon $1\Delta x$ strategy but at a similar number of samples than the $2\Delta x$ strategies, seen in Figure 9. Compared to the noiseless surface, the noisier case needs more samples to converge on a good prediction model.

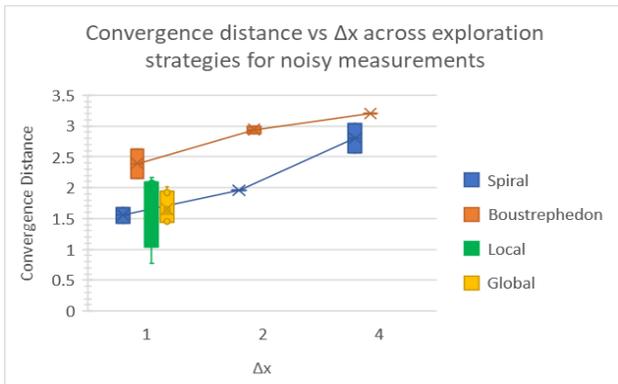

Figure 10: Comparison of distance until convergence across varying prediction and waypoint horizons on a noisy parabola

The GPAL strategies converged with less distance than the Boustrephedon strategy but similarly to the spiral strategy, Figure 10. The performance advantage is less pronounced than the noiseless case.

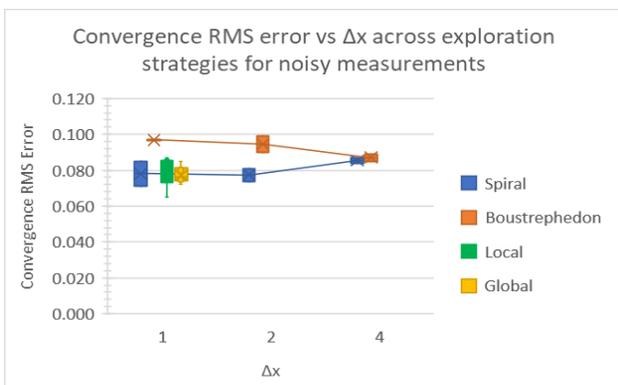

Figure 11: Comparison of RMSE until convergence across varying prediction / waypoint horizons on a noisy parabola

The GPAL strategies consistently converge to similar RMS error as the naïve strategies, Figure 11. The amount of Gaussian noise added to the true surface is $\sqrt{0.02}$ or 0.14 but the increase in error from the noiseless to noisy converged RMS error only raised 0.04, 1/3 of the added noise. The global strategy performed comparably with the local strategy.

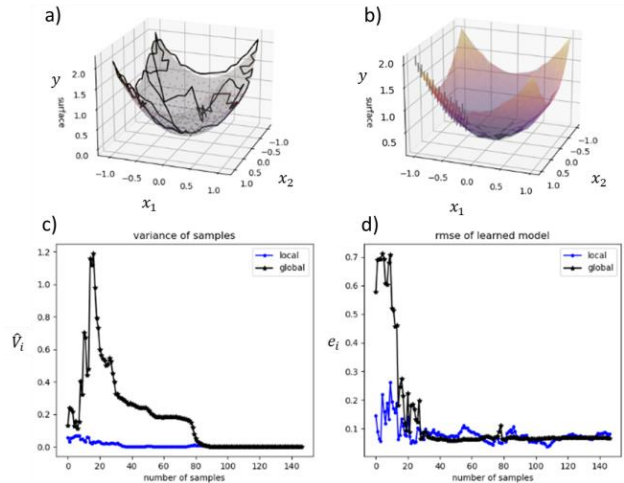

Figure 12: Example of GPAL local exploration strategy on noisy parabolic surface.

Figure 12 is the single best exploration strategy for a noisy convex surface (GPAL local). Like the noiseless parabola, the path traversed scans the edge. Variance spikes initially quickly and takes more samples to converge asymptotically, but RMS error monotonically decreases.

*Noiseless Townsend*

The Townsend benchmark surface is harder to model due to a multitude of local minima and maxima but is still smooth. The total number of samples that could be sampled is 2601. Like the noiseless parabola, the spiral and Boustrephedon strategies result in identical models (no variance in box-whisker plots) as the model inputs are identical. **Only the GPAL local exploration strategy converged in 100% of the trials**. The global GPAL exploration strategy never converged. Spiral and Boustrephedon converge in the $2\Delta x$ trial, but does so with more samples and far more distance.

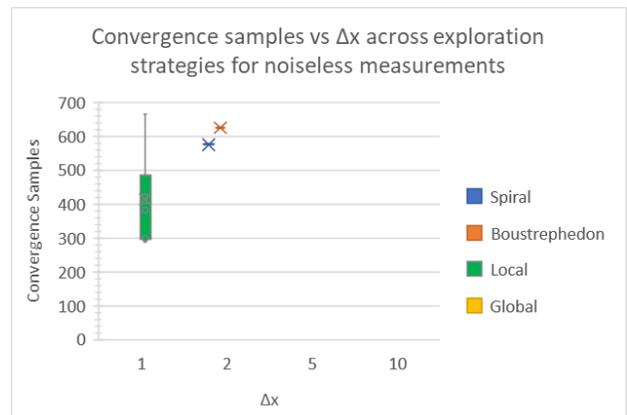

Figure 13: Comparison of samples until convergence across varying prediction and waypoint horizons on a noiseless Townsend surface. Note that trials that do not converge are gaps in the box-whisker-line plot.

GPAL local strategy converged after sampling 11 – 26% of the total samples of space, seen in Figure 13, less than all other strategies.

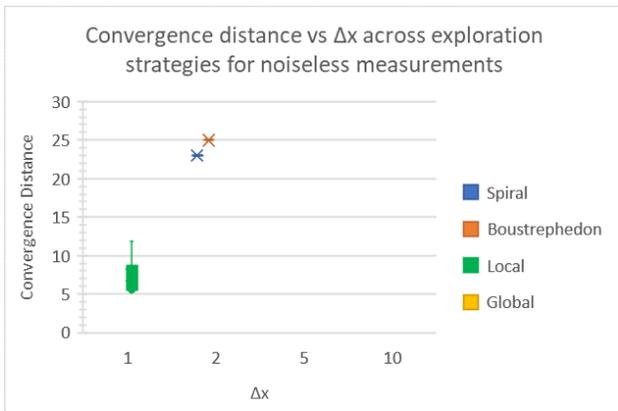

Figure 14: Comparison of distance until convergence across varying prediction/waypoint horizons on a noiseless Townsend surface

The GPAL local strategy only needed 21 – 51 % of the distance of the Spiral or Boustrephedon strategy.

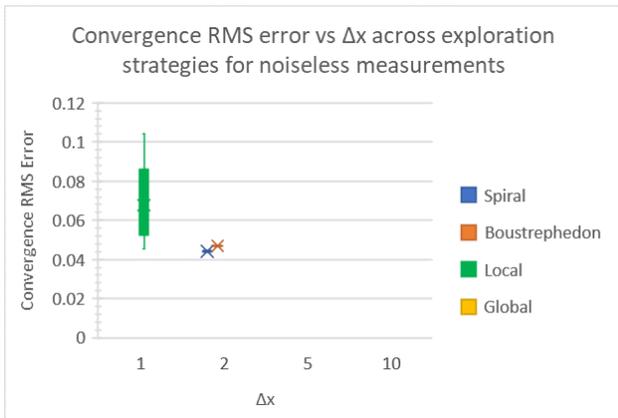

Figure 15: Comparison of RMSE until convergence across varying prediction / waypoint horizons on a noiseless Townsend surface

The converged RMS error of the local GPAL strategy is 25 – 136% more error than the Spiral and Boustrephedon strategies, Figure 15. The naïve strategies converge to a lower RMS error, which is only an advantage if you know what $\Delta x$ to sample. One can ask: Is this 3x distance worth the 2x error?

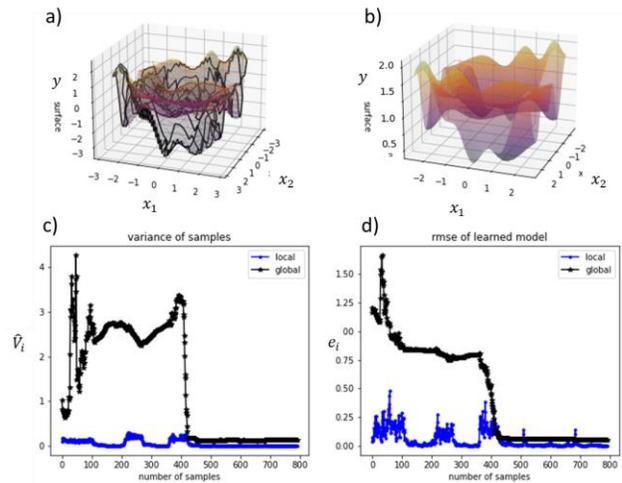

Figure 16: Example of GPAL local exploration strategy on noiseless Townsend surface.

One of the best trials of GPAL local shown in Figure 16. Given the many local wells and peaks of the Townsend surface, this surface required the vehicle to travel farther distances to cover all localities. Edges and corners are well traveled. These local wells are represented in the swelling and ebbing in the variance of samples. The variance peaks right before convergence. RMS error initially peaks but decreases suddenly just as variance does. The GPAL local strategy outperforms all other exploration strategies in minimizing prediction error. No exploration strategy can consistently identify the global minimum, which is understandable due to this benchmark surface's many local minima.

*Noisy Townsend*

Much like the noiseless trials, not many exploration strategies converge but the GPAL local strategy consistently converges 100% of the time.

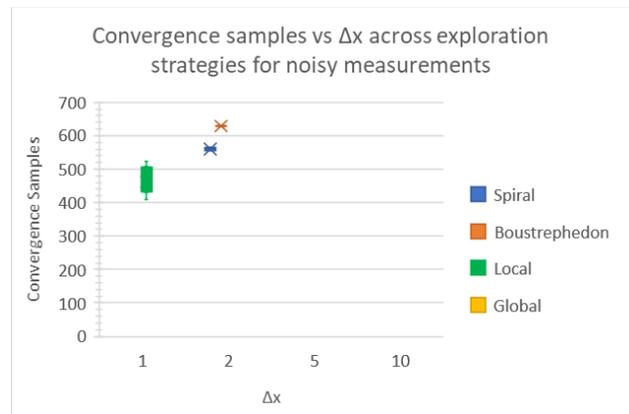

Figure 17: Comparison of samples until convergence across varying prediction and waypoint horizons on a noisy Townsend surface

Again, Spiral and Boustrephedon converge in the $2\Delta x$ trial, but does so with more samples and far more distance, seen in Figure 17 and Figure 18.

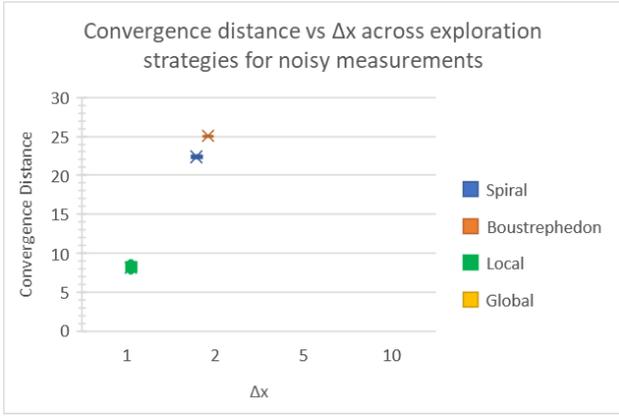

Figure 18: Comparison of distance until convergence across varying prediction and waypoint horizons on a noisy Townsend surface

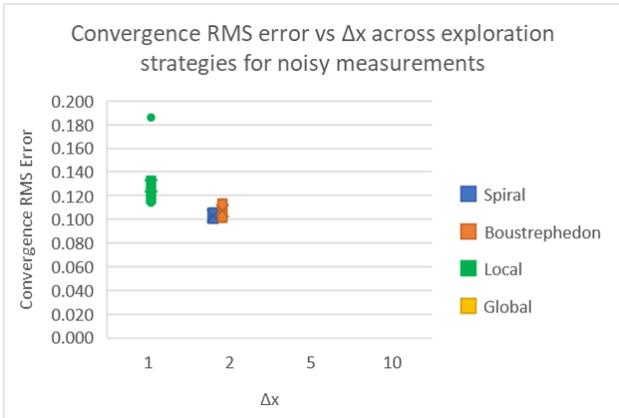

Figure 19: Comparison of RMS error until convergence across varying prediction and waypoint horizons on a noisy Townsend surface

The RMS error of GPAL is more than the Spiral RMS error, but the mean/mode is comparable, Figure 19.

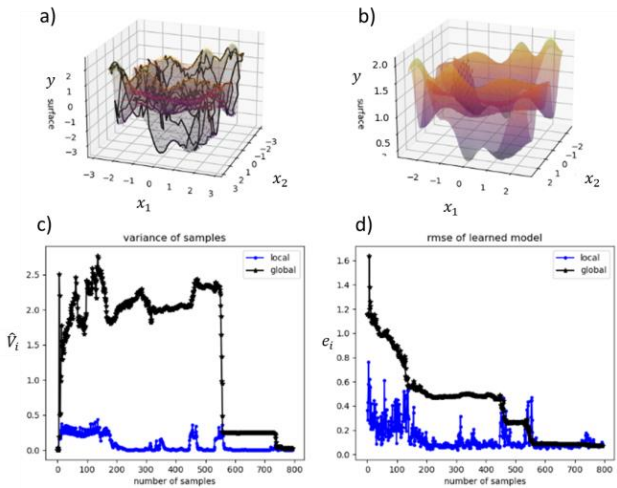

Figure 20: Example of GPAL local exploration strategy on noisy Townsend surface.

One of the best trials of the GPAL local strategy is shown in Figure 20. Again, peaks, troughs, edges, and corners are well traveled. The variance wavers and maintains a large variance until nearly sudden convergence to a suboptimal solution, then the optimal solution. RMS error decreases nearly monotonically and converges when variance converges. All exploration strategies, except for the global GPAL, moderately improve the learned model's RMS error. The addition of noise has increased the noise floor and achievable prediction error. Only the local GPAL strategy can precisely locate the global minimum.

*3 km Lunar Crater*

The lunar surface environment differs from the parabola and Townsend surfaces in that there is no true surface to compare error to. All RMS error is grounded to the observed data, which is noisy. The 3 km crater surface also has 3 times less data (169 samples) than the smallest benchmark surface, the parabola. For the lunar experiments, a prediction horizon resolution was added: $1\Delta x$, which is analogous to the nearest neighbors (NN).

Of the GPAL strategies, the local strategy surprisingly does the slightly worse in all metrics (samples, distance, and RMS error), global strategy middle of the pack, and NN strategy the best. The global strategy does not always converge. The local and NN methods always settle into a plateau of RMS error. Interestingly, the only science-blind method to converge is the spiral method for fine resolution maps ($1\Delta x$ and $2\Delta x$). Despite more moderate RMS error improvements made by the GPAL strategies, the GPAL strategies can consistently identify the global minimum location, whereas the science-blind strategies cannot.

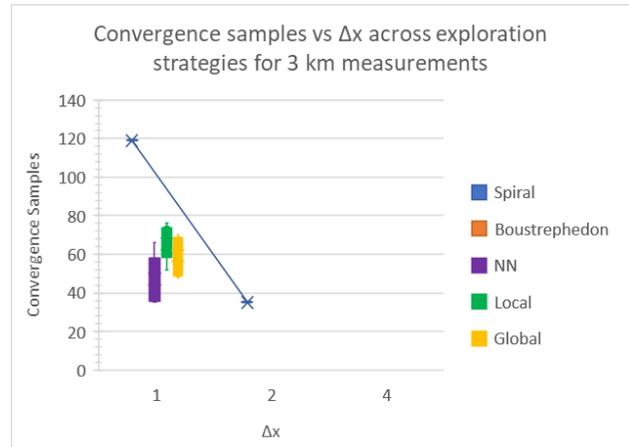

Figure 21: Comparison of samples until convergence across varying prediction and waypoint horizons on a noisy 3 km lunar crater surface

Of the total 169 samples, the NN strategy requires 35 – 66 samples (20% - 39%) to converge. The local strategy needs 52 – 76 samples (31% - 45%) to. The global strategy requires 48 – 70 samples (28% - 41%) to converge.

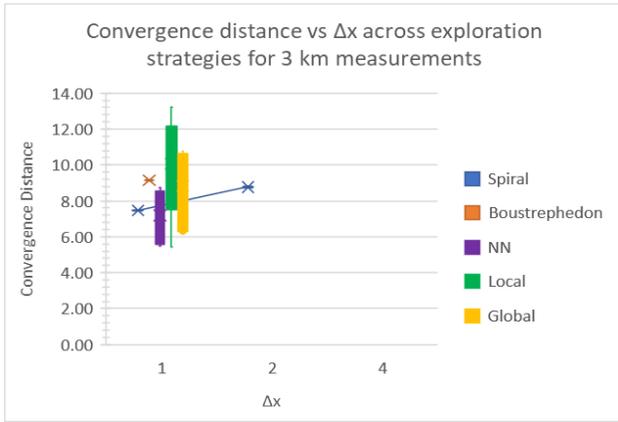

Figure 22: Comparison of distance until convergence across varying prediction and waypoint horizons on a noisy 3 km lunar crater surface

For distance, the NN horizon performed significantly better than the local strategy, reducing the distance necessary by (17% - 54%). The NN performed similarly in distance with the spiral method, but how does one decide what resolution to deploy a science-blind spiral strategy? Spiral also does not offer any confidence towards convergence. The NN strategy converged to RMS error values lower than the local strategy.

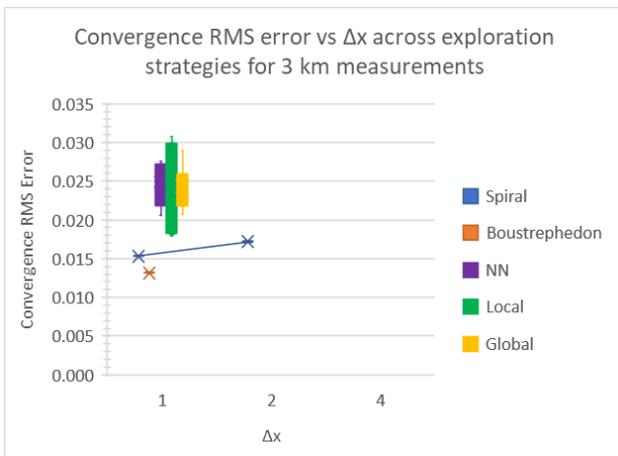

Figure 23: Comparison of RMS error until convergence across varying prediction and waypoint horizons on a noisy 3 km lunar crater surface

The spiral method does perform competitively: the trivial $1\Delta x$ method does necessitate the most samples (Figure 21), but achieves the best RMS error (Figure 23) and does so with less distance than most GPAL runs (Figure 22).

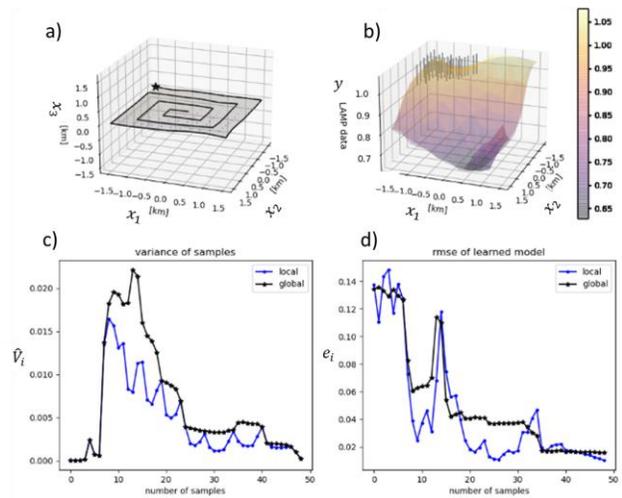

Figure 24: Example of spiral $2\Delta x$ exploration strategy on noisy lunar crater surface.

The $2\Delta x$ spiral method, seen in Figure 24, requires the least amount of sample for trials that converged, middle of the pack in distance, and still achieves better RMS error than all GPAL runs.

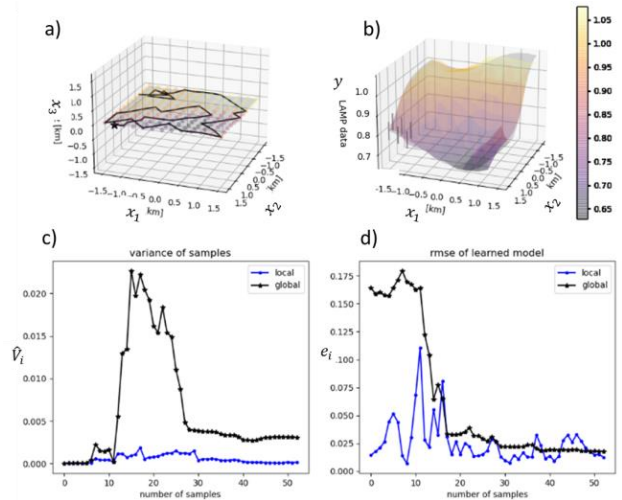

Figure 25: Example of GPAL NN exploration strategy on noisy 3 km lunar crater surface.

Generally, the NN strategy travels to the edges quite quickly and revisits spots rarely, as can be seen in Figure 25. The variance peaks early on and settles. The RMS error oscillates, then monotonically decreases, converges.

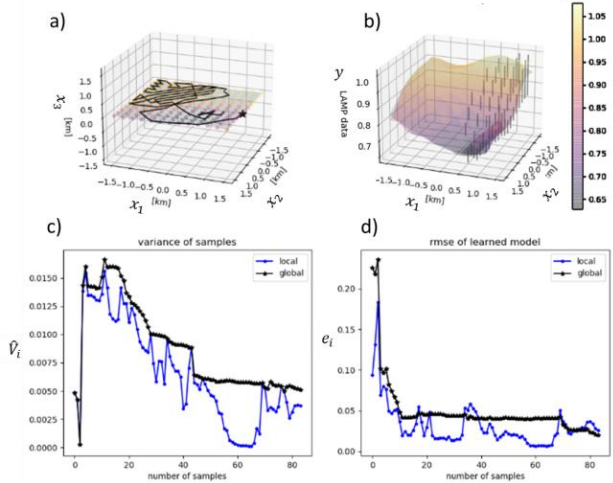

Figure 26: Example of GPAL local strategy on a noisy 3 km lunar crater surface.

The local strategy travels to one corner and passes over the same area, as can be seen in Figure 26. The other corners were not visited. The local strategy could have included too much surface area in its evaluation of where to go next.

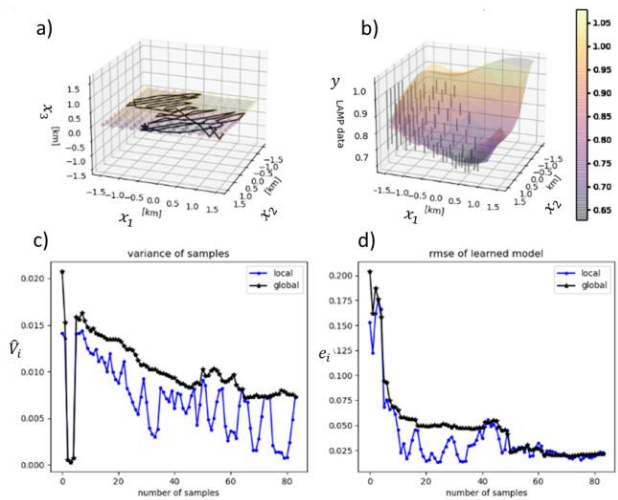

Figure 27: Example of GPAL global strategy on a noisy 3 km lunar crater surface.

The global strategy is even more so distracted and zips back and forth, as can be seen in Figure 27. Somehow, the corner predictions still resemble the topography. The variance decreases much more slowly and steadily. The RMS error slowly decreases but settles.

*6 km Lunar Crater*

This 6 km surface differed from parabola and Townsend surfaces in that the lunar crater surface was not smooth. The peaks are perceivably sharper. The total number of samples that could be sampled is 625. All GPAL strategies guaranteed convergence. The NN strategy outperformed the other strategies in sampling and distance. GPAL strategies show more moderate RMS error improvements made by the and exact prediction of global minima for the 6 km case.

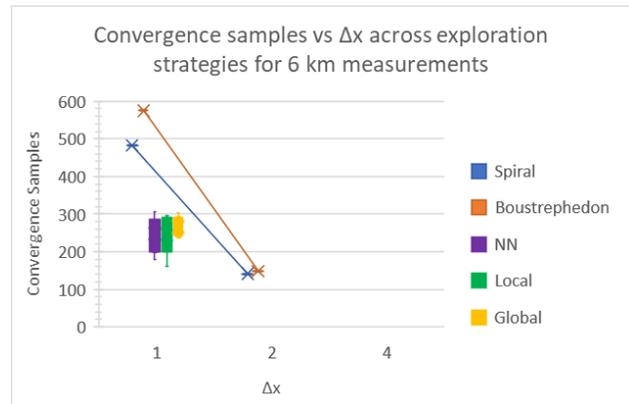

Figure 28: Comparison of samples until convergence across varying prediction and waypoint horizons on a noisy 6 km lunar crater surface

The GPAL strategies needed less samples to converge than the naïve $1\Delta x$ strategies but a bit more than the $2\Delta x$ strategies, Figure 28.

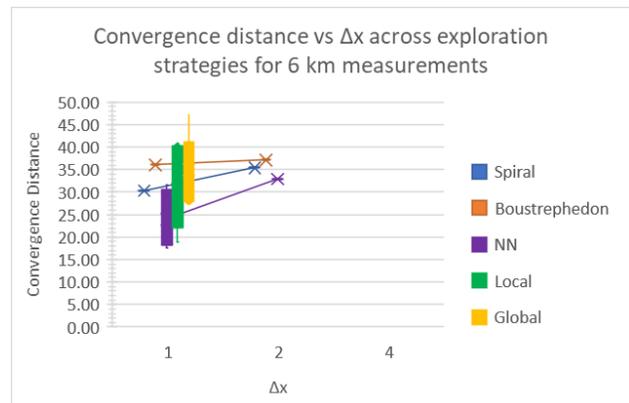

Figure 29: Comparison of distance until convergence across varying prediction and waypoint horizons on a noisy 6 km lunar crater surface

The convergence distance performance decreased from global to local to NN, Figure 29.

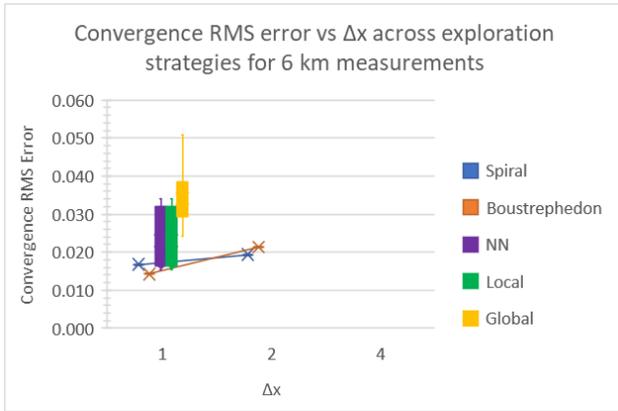

Figure 30: Comparison of RMS error until convergence across varying prediction and waypoint horizons on a noisy 6 km lunar crater surface

The RMS error showed the NN strategy converged on the least amount of error, Figure 30. The global strategy converged on the most error and the local strategy more than the NN error.

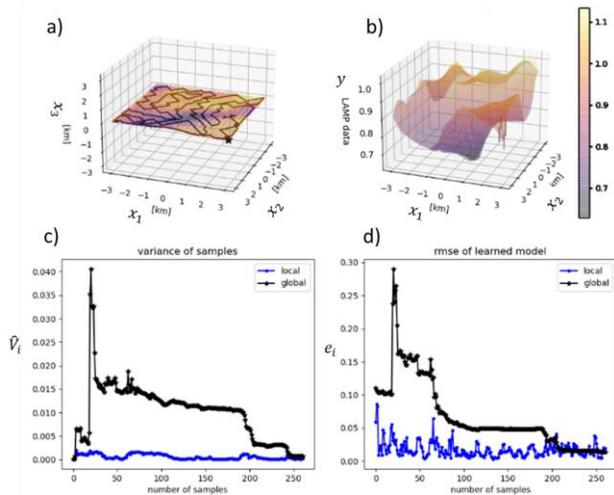

Figure 31: Example of GPAL NN exploration strategy on noisy 6 km lunar crater surface.

The nearest neighbor strategy shows a path that prioritizes edges and corners and almost distributes coverages uniformly across the entire surface, Figure 31. The variance peaks early and decreases nearly monotonically. The RMS error follows a similar pattern.

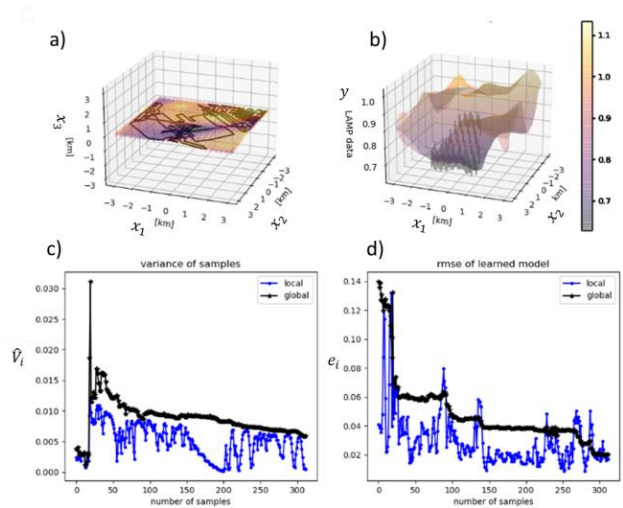

Figure 32: Example of GPAL local exploration strategy on noisy 6 km lunar crater surface.

The local strategy prioritizes coverage of three locations, the highest peak, the lowest valley, and an edge, seen in Figure 32. The oversight in the other corners may be that the algorithm was tugged in many different directions, causing paralysis/stuttering. The RMS error monotonically decreases in step functions and is thought to converge, although the variance does not drop in a step function, typical of other more notable convergence. The global error is within local error, so that is a good sign of convergence.

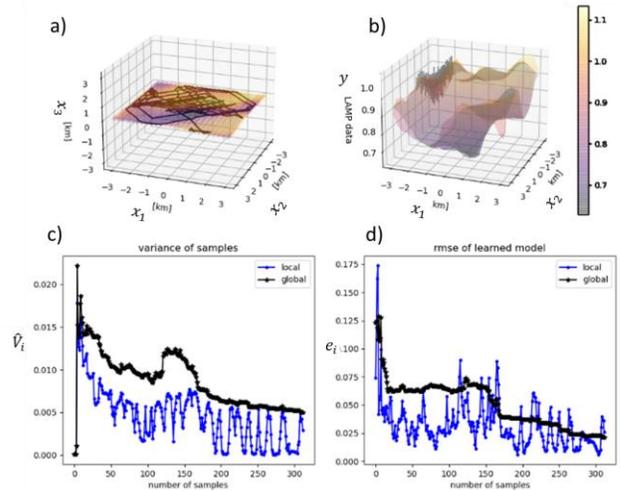

Figure 33: Example of GPAL global exploration strategy on noisy 6 km lunar crater surface.

The global strategy sweeps across and revisits the middle section of the environment, missing some finer peaks and valleys, Figure 33. The variance spikes initially, with the global variance staying above the local variance throughout the entire trial. The RMS error monotonically decreases and the global error is within local error. Global variance doesn't necessarily need to match local variance for global RMS error to match local RMS error, although an analysis of variance is planned in future work.

## 7. CONCLUSIONS

This paper investigates the efficacy of science-blind and information-driven exploration strategies in comprehensively characterizing and predicting the scalar distribution across a surface. The science-blind methods, which include the spiral and Boustrephedon, are effective in understanding simple surfaces with no noise, but are less likely to converge when surfaces become more complex and noisier. **By and large, the greedy GPAL exploration strategies offered more guarantees of RMS error convergence, comprehensive coverage, and consistent global minimum identification**.

An interesting observation after running 82 total experiments, half of which were **Gaussian process informed strategies**, is that **the smallest prediction and waypoint horizons yielded the most comprehensive surface coverage, which led to faster convergence, shorter traverse distances, and better model predictions.** There is a conversation to be had about how myopia or short-sightedness is beneficial to an exploration strategy in which the most greedy or reactive actions are most rewarding.

This body of work is a machine learning practitioner's empirical observation that have raised the following questions that will inspire future work:

1) How does model prediction (posterior) error relate to model covariance and Shannon entropy?
2) Can model error converge without a true reference model?
3) What is the impact of a finite horizon when dictating the next waypoint as a policy hyperparameter on minimizing learned model error and distance traveled?

## BIOGRAPHY

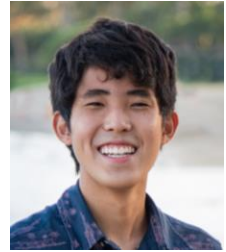

*Ashten Akemoto is currently pursuing a B.S. in Computer Engineering at the University of Hawaii at Manoa. Ashten is currently a Hawaii Space Grant Consortium Research Intern and Software Engineer for the Robotic Space Exploration VIP team. Ashten is studying to eventually do software engineering work in the field of autonomous space exploration.*

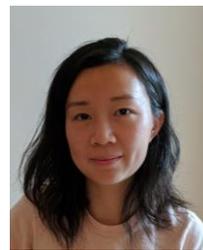

*Frances Zhu earned her B.S. in Mechanical and Aerospace Engineering from Cornell University, Ithaca in 2014 and a Ph.D. in Aerospace Engineering at Cornell in 2019. Dr. Zhu was a NASA Space Technology Research Fellow. Since 2020, she has been an assistant research professor with the Hawaii Institute of Geophysics and Planetology at University of Hawaii, specializing in machine learning, dynamics, systems, and controls engineering. She is also the deputy director for the Hawaii Space Grant Consortium and graduate cooperating faculty with the following departments: mechanical engineering, electrical engineering, and earth science.*